\documentclass[11pt]{article}

\usepackage[pdftex]{graphicx}
\usepackage{hyperref}
\usepackage{amssymb}
\usepackage{amsmath}
\usepackage{bm}
\usepackage{mathtools}
\usepackage{booktabs}
\usepackage{algorithm}
\usepackage[noend]{algorithmic}

\title{\vspace*{-0.75in}\bf Creative AI Through Evolutionary
  Computation: Principles and Examples}
\author{Risto Miikkulainen\\
  The University of Texas at Austin and Cognizant Technology Solutions}
\small\normalsize
\date{}
\begin{document}
\maketitle

\begin{abstract}
The main power of artificial intelligence is not in modeling what we
already know, but in creating solutions that are new. Such solutions
exist in extremely large, high-dimensional, and complex search
spaces. Population-based search techniques, i.e.\ variants of
evolutionary computation, are well suited to finding them. These
techniques make it possible to find creative solutions to practical
problems in the real world, making creative AI through evolutionary
computation the likely ``next deep learning.''
\end{abstract}

\section{Introduction}

In the last decade or so we have seen tremendous progress in
Artificial Intelligence (AI). AI is now in the real world, powering
applications that have a large practical impact. Most of it is based
on modeling, i.e.\ machine learning of statistical models that make it
possible to predict what the right decision might be in future
situations. For example, we now have object recognition, speech
recognition, game playing, language understanding, and machine
translation systems that rival human performance, and in many cases
exceed it
\cite{awadalla:parity18,hessel:arxiv17,russakovsky:arxiv14}. In each
of these cases, massive amounts of supervised data exist, specifying
the right answer to each input case. With current computational
capabilities, it is possible to train neural
networks to take advantage of the data. Therefore, AI works great in
tasks where we already know what needs to be done.

The next step for AI is machine creativity. Beyond modeling there is a
large number of tasks where the correct, or even good, solutions are
not known, but need to be discovered. For instance designing
engineering solutions that perform well at low costs, or web pages
that serve the users well, or even growth recipes for agriculture in
controlled greenhouses are all tasks where human expertise is scarce
and good solutions difficult to come by
\cite{dupuis:ijss15,hu:aiedam08,ishida:bullet18,johnson:plos19,miikkulainen:iaai18}. Methods
for machine creativity have existed for decades. I believe we are now
in a similar situation as deep learning was a few years ago: with the
million-fold increase in computational power, those methods can now be
used to scale up to real-world tasks.

This paper first identifies challenges in creative tasks, suggests how
evolutionary computation may be able to solve them, and reviews three
practical examples of Creative AI through Evolutionary Computation.

\section{Challenges in Machine Creativity}

\begin{figure}
  \begin{center}
    \begin{minipage}{0.5\textwidth}
      \centering
      \includegraphics[width=\textwidth]{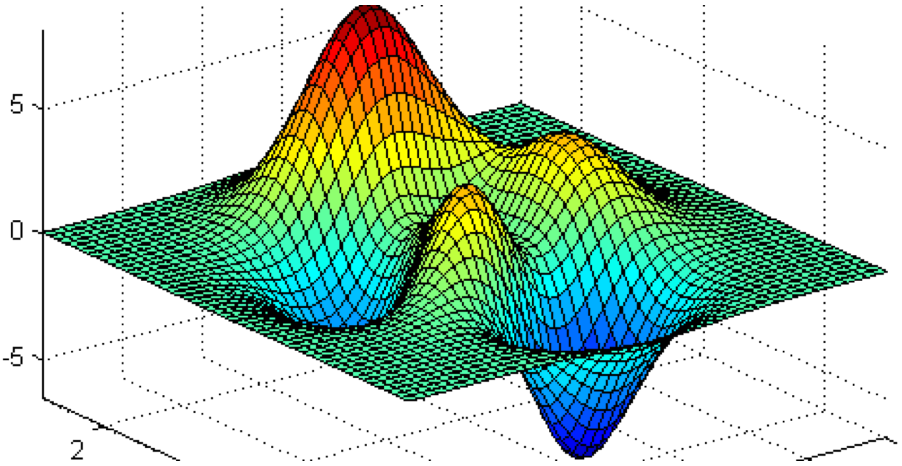}\\
      {\footnotesize (a) Search Space Appropriate for Hill Climbing}
    \end{minipage}
    \hfill
    \begin{minipage}{0.4\textwidth}
      \centering
      \includegraphics[width=\textwidth]{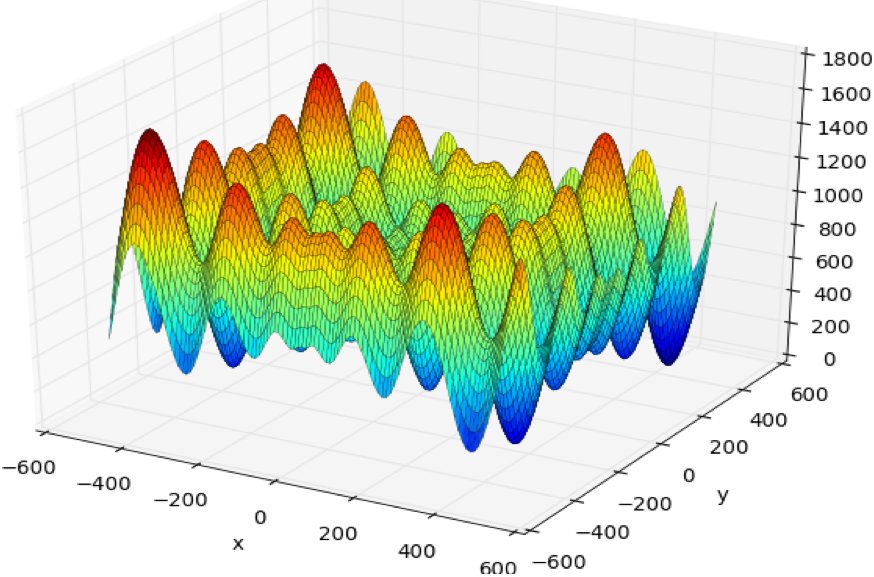}\\
      {\footnotesize (b) Search Space in a Creative Domain}
    \end{minipage}      
    \caption{Challenge of Creative Problem Solving. Human design
      process as well as deep learning and reinforcement learning can
      be seen as hill-climbing processes. They work well as long as
      the search space is relatively small, low-dimensional, and well
      behaved. However, creative problems where solutions are not
      known may require search in a large, high-dimensional space
      with many local optima. Population-based search through
      evolutionary computation is well-suited for such problems: it
      discovers and utilizes partial solutions, searches along
      multiple objectives, and novelty. (Image credit:
      http://deap.readthedocs.io/en/latest/api/benchmarks.html)}
    \label{fg:hillclimb}
  \end{center}
\end{figure}

Evolutionary computation is in a unique position to take advantage of
that power, and become the next deep learning. To see why, let us
consider how humans tackle a creative task, such as engineering
design. A typical process starts with an existing design, perhaps an
earlier one that needs to be improved or extended, or a design for a
related task. The designer then makes changes to this solution and
evaluates them. S/he keeps those changes that work well and discards
those that do not, and iterates. The process terminates when a desired level of
performance is met, or when no better solutions can be found---at
which point the process may be started again from a different initial
solution. Such a process can be described as a hill-climbing process
(Figure~\ref{fg:hillclimb}$a$). With good initial insight it is possible
to find good solutions, but much of the space remains unexplored and
many good solutions may be missed.

Interestingly, current machine learning methods are also based on
hill climbing. Neural networks and deep learning follow a gradient
that is computed based on known examples of desired behavior
\cite{lecun:nature15,schmidhuber:nn15}.  The gradient specifies how
the neural network should be adjusted to make it perform slightly
better, but it also does not have a global view of the landscape,
i.e.\ where to start and which hill to climb. Similarly, reinforcement
learning starts with an individual solution and then explores
modifications around that solution, in order to estimate the gradient
\cite{salimans:arxiv17,zhang:arxiv17}. With large enough networks and
datasets and computing power, these methods have achieved remarkable
successes in recent years.

However, the search landscape in creative tasks is likely to be less
amenable to hill climbing (Figure~\ref{fg:hillclimb}$b$). There are
three challenges: (1) The space is large, consisting of too many
possible solutions to be explored fully, even with multiple restarts;
(2) the space is high-dimensional, requiring that good values are
found for many variables at once; and (3) the space is deceptive,
consisting of multiple peaks and valleys, making it difficult to make
progress through local search.

\section{Evolutionary Computation Solution}

Evolutionary computation, as a population-based search technique, is
in a unique position to meet these challenges. First, it makes it
possible to explore many areas of the search space at once. In effect,
evolution performs multiple parallel searches, not a single
hill climb. By itself such parallel search would result in only a
linear improvement, however, the main advantage is that the searches
interact: if there is a good partial solution found in one of the
searches, the others can immediately take advantage of it as
well. That is, evolution finds building blocks, or schemata, or
stepping stones, that are then combined to form better comprehensive
solutions \cite{forrest:foga93,holland:adaptation,meyerson:gecco17}.

This approach can be highly effective, as shown e.g.\ in the benchmark
problem of multiplexer design\cite{koza:foga91}.  Multiplexers are
easy to design algorithmically: the task is to output the bit (among
$2^n$ choices) specified by an $n$-bit address. However, when
formulated as a search problem in the space of logical operations this
problem is challenging because the search space grows very quickly,
i.e.\ as $2^{2^{n+2^n}}$.  There is, however, structure in that space
that evolution can discover and utilize effectively. It turns out that
evolution can discover solutions in extremely large cases, including
the 70-bit multiplexer (i.e.\ $n=6$) with a search space of at least
$2^{2^{70}}$ states. It is hard to conceptualize a number that large,
but to give an idea, imagine having the number printed using a 10pt font
on a piece of paper. It would take light 95 years to traverse from the
beginning to the end of that number.

Second, population-based search makes it possible to find solutions in
extremely high-dimensional search spaces as well. Whereas it is very
difficult to build a model with high-order interactions beyond pairs
or triples, the population represents such interactions implicitly, as
the collection of actual combinations of values that exist in the good
solutions in the population. Recombination of those solutions then
makes it possible to collect good values for a large number of
dimensions at once.

As an example, consider the problem of designing an optimal schedule
for metal casting \cite{deb:ejor17}. There are variables for number of
each type of object to be made in each heat (i.e.\ melting
process). The number of objects and heats can be grown from a few
dozen, which can be solved with standard methods, to tens of
thousands, resulting in billion variables. Yet, utilizing an
initialization process and operators customized to exploit the
structure in the problem, it is possible to find good combinations for
them, i.e.\ find near-optimal solutions in a billion-dimensional
space. Given that most search and optimization methods are limited to
several orders of magnitude fewer variables, this scaleup makes it
possible to apply optimization to entire new category of problems.

Third, population-based search can be adapted naturally to problems
that are highly deceptive. One approach is to utilize multiple
objectives \cite{deb:ppsn00}: if search gets stuck in one dimension,
it is possible to make progress among other dimensions, and thereby
get around deception. Another approach is to emphasize novelty, or
diversity, of solutions in search \cite{stanley:book15}. The search
does not simply try to maximize fitness, but also favors solutions
that are different from those that already exist. Novelty can be
expressed as part of fitness, or a separate objective, or serve as a
minimum criterion for selection, or as a criterion for mate selection
and survival \cite{cuccu:evostar11,gomes:gecco15,lehman:gecco10a,mcquesten:phd02,mouret:evolcomp12}.

For instance, in the composite novelty method \cite{shahrzad:alife18},
different objectives are defined for different aspects of performance,
and combined so that they specify an area of search space with useful
tradeoffs. Novelty is then used as the basis for selection and
survival within this area. This method was illustrated in the problem
of designing minimal sorting networks, which have to sort a set of $n$
numbers correctly, but also consist of as few comparator elements as
possible (which swap two numbers), and as few layers as possible
(where comparisons can be performed in parallel). The search space is
highly deceptive because often the network structure needs to be
changed substantially to make it smaller. Combining multiple
objectives and novelty finds solutions faster and finds better
solutions than traditional evolution, multiobjective evolution, and
novelty search alone. The approach already found a new minimal network
for 20 inputs \cite{shahrzad:gptp20}.

Thus Evolutionary Computation has the right properties to solve
challenging tasks that require creativity. The next three subsections
review three examples on how this power can be put to use in
discovering creative solutions in real-world applications.

\section{Designing Effective Web Interfaces}

The first example is Ascend by Evolv, an actual commercial application
of evolutionary computation on conversion optimization, i.e.\ on
designing web interfaces to make it more likely that a user will take
the desired action on the page, such as signing them up, buying
something, or requesting for more information
\cite{miikkulainen:aimag20,miikkulainen:gecco17ascend}.

More specifically, the human expert defines a search space, consisting
of a set of elements on the page, such as the heading text, size, and
color, background image, and content order, possible values for each,
and possible restrictions among their combinations. Page design can
then be represented as a vector and optimized using genetic
algorithms. Each candidate is evaluated by deploying it on the web.  A
sufficient number of actual users is directed to each candidate
design, and how well they convert is measured.

Typically about 2000 users are needed to estimate a typical 1-4\%
conversion rate for evolution to make progress. With a population of a
few dozen candidates, Ascend usually discovers designs that are better
than control in 10-20 generations. Ascend has been applied to hundreds
of web interfaces across a variety of industries and search space
sizes, and it routinely improves performance 10-200\% over the
original human designs.

\begin{figure}
  \begin{center}
    \includegraphics[width=0.6\textwidth]{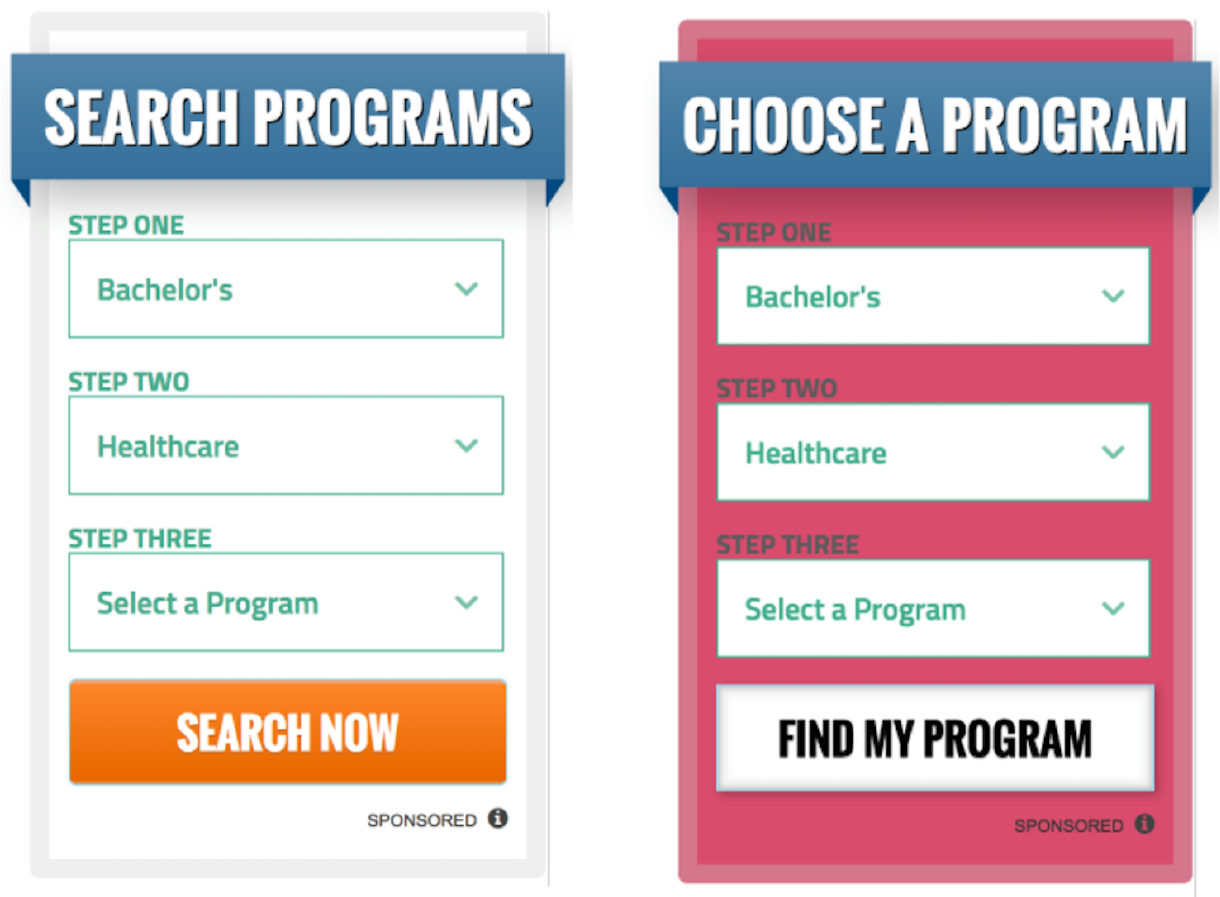}\\
    \hfill\hfill(a) Control\hfill\hfill(b) Evolved\hfill\hfill\mbox{}\\
    \caption{A comparison of human design and evolutionary design for a
      sign-up widget in web design. (a) The original design is clear and
      consistent, according to general design principles. (b) The
      evolutionary design is brash and bold, and unlikely to be designed by
      humans. However, it converts 45\% better, demonstrating that
      evolution can discover creative solutions that humans miss.}
    \label{fg:ascend}
  \end{center}
\end{figure}

However, what is most interesting about Ascend is that it can discover
creative solutions that human designers miss. The humans utilize
principles of perceptual psychology and aesthetics, such as hierarchy,
directionality, consistency, and clarity, but it turns out following
them does not necessarily make the page effective. An example is shown
in Figure~\ref{fg:ascend}. While the control design is elegant, the
design discovered by evolution is brash, using neon colors, contrast,
and strong text. As a matter of fact, when evolution was running, it
came up with similar designs so frequently that the designers labeled
it ``the ugly widget generator.'' However, this ugly widget performs
45\% better than the control! This result suggests that there is still
much that we do not know about factors that affect conversions, but
evolution can nevertheless learn them and utilize them in creative
design.

\section{Discovering Growth Recipes for Agriculture}

The second example takes advantage of surrogate modeling, a powerful
extension that makes it possible to apply evolutionary creativity to
many more problems in the real world. Whereas the Ascend designs could
be evaluated in the real world with little cost, in many other
domains, such as healthcare, finance, or mechanical design, such
evaluations, especially of the most creative solutions, could be
costly or dangerous.

A case in point is developing growth recipes for computer-controlled
agriculture. That is, given vertical farming environments where the
inputs such as water, temperature, nutrients, and light can be
controlled at will, the challenge is to determine how those inputs
should be set so that the plants grow as well as possible, for
instance improving flavor, maximizing size, or minimizing
cost. Whereas it takes a long time to grow a plant, a large number of
recipes can be evaluated immediately with a surrogate model.

In a pioneering experiment, recipes were developed for
optimizing flavor in basil, focusing on light variables such as
wavelength, period, and UV component. Initially a few hundred recipes
were implemented in real growth containers, representing known good
recipes as well as recipes that covered the space more broadly. A
surrogate model was trained on the resulting data, with flavor
measured in terms of volatile composition of the plants. About
a million recipes were then created through search and evaluated
against the surrogate. In the end, the best ones were evaluated in
real growth experiments.

\begin{figure}
  \begin{center}
    \includegraphics[width=0.6\textwidth]{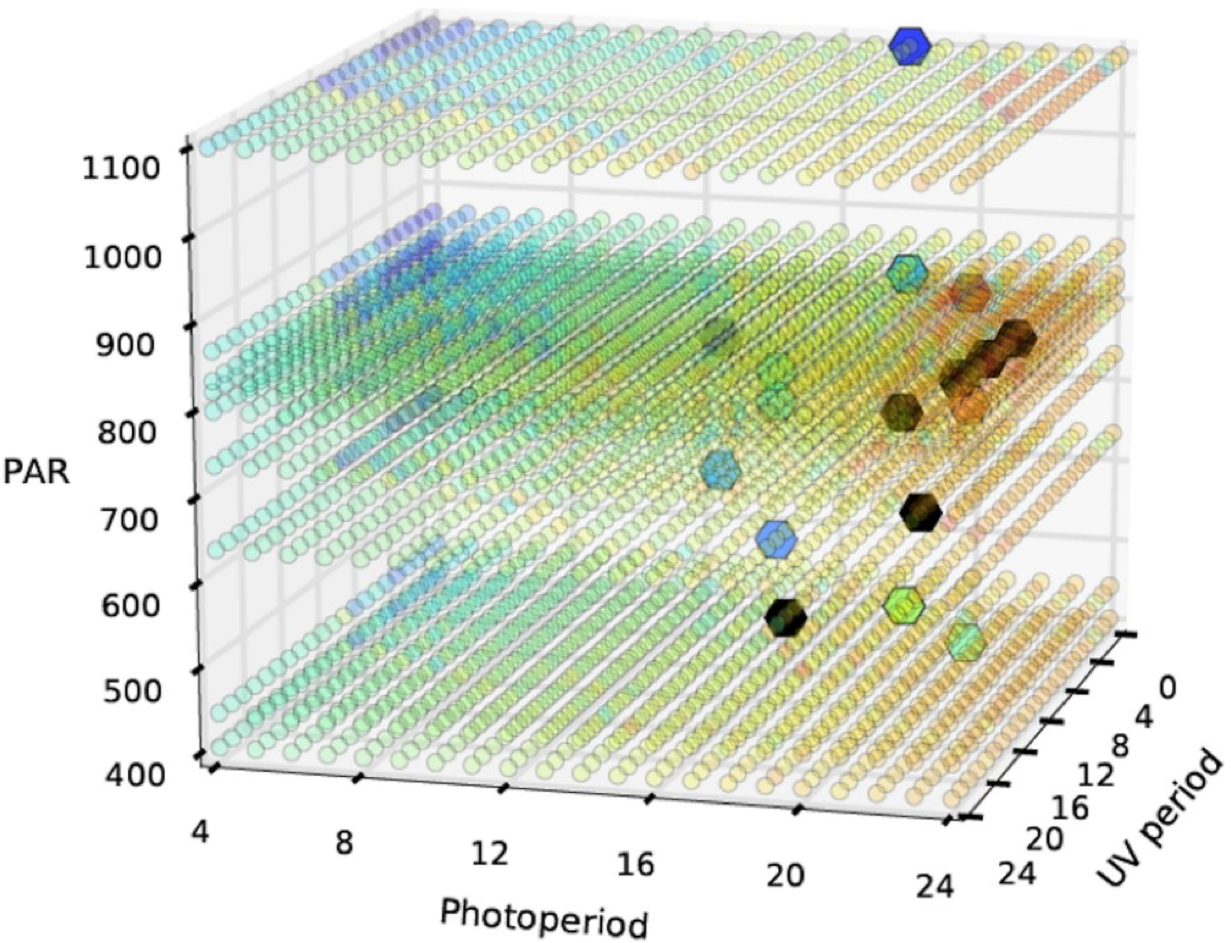}\\
    \caption{Discovering a counterintuitive 24-hr light period for
      computer-controlled agriculture. With the initial 18-hr
      restriction removed, evolution discovered that when the lights
      are always on, basil will develop more flavor. The axes
      represent the three light variables, with light period on the
      horizontal axis. The color of the small dots indicates their
      value predicted by the model (red $>$ yellow $>$ green
      $>$ blue). The large dots are suggestions, and the darker dots are
      the most recent ones. In this manner, if given a search space
      free of human biases, evolution can discover effective,
      surprising solutions.
    }
    \label{fg:cyberag}
  \end{center}
\end{figure}

In this process, a most remarkable discovery was made.
Initially the maximum light period was set to 18 hours, assuming that
the daily light cycle in the real world was a reasonable constraint.
However, search quickly discovered that recipes at 18 hours were
the best. At that point, the daily cycle restriction was removed---and
(as shown in Figure~\ref{fg:cyberag}), even better recipes were found
with a light period of 24 hours!  This result is counterintuitive and
it was a surprise to the biologists in the team. It demonstrates how
human biases can get in the way of discovering good solutions.
Evolution does not have such biases, and if given enough freedom to
explore, can create effective, surprising solutions.

\section{Finding Mitigation Strategies for COVID-19}

The third example is particularly topical at the time of this writing:
Determining how various countries could implement non-pharmaceutical
interventions (NPIs), such as school and workplace closings,
restrictions on gatherings and events, and limitations on movement, in
order to reduce the spread of the pandemic with minimal economic cost
\cite{miikkulainen:arxiv20}. It is also significantly more complex in
that, in addition to requiring a surrogate for evaluation, the
solutions are strategies, represented by neural networks, instead of
single points (such as web-page designs or growth recipes).

More specifically, the approach consists of first training a
Predictor, i.e. a surrogate model, to predict how the number of cases
would develop in the future, given a history of cases and NPIs in a
country in the past, and an NPI strategy for the future. Using
historical NPI data from the Oxford COVID-19 government response
tracker and case data from Johns Hopkins COVID-19 Data Repository, it
was possible to train a recurrent LSTM neural network for this
purpose. As opposed to traditional epidemiological models, such a
model is purely phenomenological, includes all hidden interactions,
and turned out to be surprisingly accurate, even given that the data
is collected and used as the pandemic unfolds.

\begin{figure}
  \begin{center}
    \includegraphics[width=0.8\textwidth]{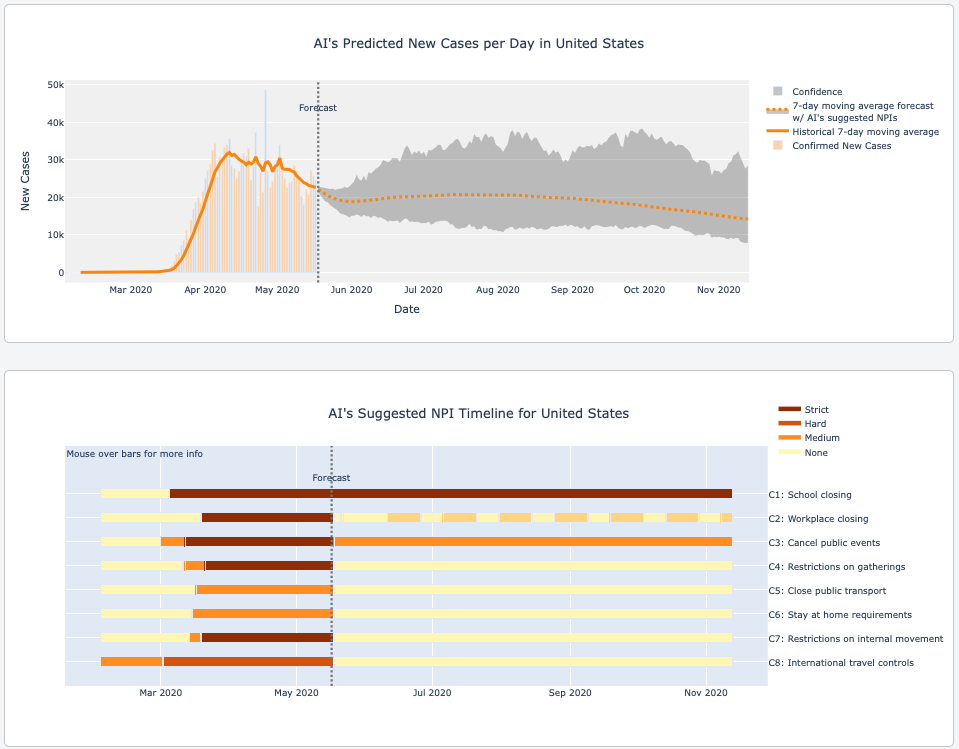}\\
    \caption{An example creative solution for opening the economy
      after the COVID-19 peak had passed. The top plot shows the
      historical past and predicted future number of cases in the US
      on May 18th, 2020.  The bottom plot illustrates the NPIs in
      effect or recommended during the same timeline, with color
      coding indicating their stringency. The system still recommends
      restrictions on schools, workplaces, and public events (top
      three rows), but suggests that opening and closing workplaces
      can be alternated, thus mitigating the effect on both economy
      and cases.}
    \label{fg:npi}
  \end{center}
\end{figure}

In the second step, Prescriptor neural networks, representing the NPI
strategies, were then evolved, using the Predictor as a surrogate to
evaluate how effective they were. Since there are two conflicting
objectives (minimize cases vs. stringency of NPIs), the result is a
Pareto front that trades off these objectives: Some prescriptors
keep the number of cases down by locking down, others keep the society
open with the cost of more cases. The decision maker can then select a
desired tradeoff, and the Prescriptor will recommend the best NPI
strategy that achieves it.

This process, Evolutionary Surrogate-assisted Prescription (ESP;
\cite{francon:gecco20}), made several creative discoveries. Early on it
recognized that schools and workplaces are the most important NPIs;
Indeed these are the two activities where people spend a lot of time
with other people indoors, where it is possible to be exposed to
significant amounts of the virus, as later became evident
\cite{kay:quillette20,lu:emerindis20,park:emerindis20}. After the peak
has passed and economies are opening up, it discovered that
alternating between opening and closing schools and workplaces could
be an effective way to lessen the impact on the economy while reducing
cases (Figure~\ref{fg:npi}). While it may sound unwieldy, it has
recently been suggested as a possibility
\cite{chowdhury:eurjourepi20}.  Given the limited search space
available for evolution, it is a creative solution for lifting the
NPIs gradually. Coming out of a lockdown, it recognized that people
are less likely to adher to restrictions than they were going in, and
therefore recommended more NPIs that can be enforced, such as
restrictions on events and international travel. In this manner, the
data-based modeling and evolutionary discovery was able to track the
changing context of the pandemic, and recommend creative new
responses. Counterfactual studies with past data suggested that they
could have indeed been more effective than the actual NPIs implemented
at the time \cite{miikkulainen:arxiv20}. For an interactive demo of
the system, see \url{https://evolution.ml/esp/npi}.

\section{Conclusion}

To conclude, evolutionary computation is an AI technology that is on
the verge of a breakthrough, as a way to take machine creativity to
the real world. Like deep learning, it can take advantage of
computational resources that are now becoming available. Because it is a
population-based search method, it can scale with compute better than
other machine learning approaches, which are largely based on
hill climbing. With evolution, we should see many applications in the
near future where human creativity is augmented by evolutionary search
in discovering complex solutions, such as those in engineering,
healthcare, agriculture, financial technology, biotechnology, and
e-commerce, resulting in more complex and more powerful solutions than
are currently possible.

\bibliographystyle{myplain}
\bibliography{/u/nn/bibs/nnstrings,/u/nn/bibs/nn,/u/risto/risto}
\end{document}